\documentclass{article}
\usepackage{spconf,amsmath,graphicx}
\usepackage{graphicx}
\usepackage{amsmath}
\usepackage{amssymb}
\usepackage{booktabs}
\usepackage{times}
\usepackage{soul}
\usepackage{url}
\usepackage[colorlinks,linkcolor=black,anchorcolor=black,citecolor=black]{hyperref}
\usepackage[utf8]{inputenc}
\usepackage{amsmath}
\usepackage{amsthm}
\usepackage{amsfonts}
\usepackage{booktabs}
\usepackage{algorithm}
\usepackage{algorithmic}
\usepackage[switch]{lineno}
\usepackage{color}
\usepackage{multirow}
\usepackage{float}
\usepackage{colortbl}
\usepackage[sort,numbers]{natbib} 

\title{ARFA: An Asymmetric Receptive Field Autoencoder Model for Spatiotemporal Prediction}
\name{Wenxuan Zhang$^{1, }$$^*$, Xuechao Zou$^{1, }$$^*$, Li Wu$^{1, }$$^\dagger$\thanks{$^*$These authors contributed equally.} \thanks{$^\dagger$Corresponding author.}, Xiaoying Wang$^{1}$, Jianqiang Huang$^{1}$, Junliang Xing$^{2}$}
\address{
$^1$Qinghai University, Department of Computer Technology and Applications, Xining, China\\
$^2$Tsinghua University, Department of Computer Science  and Technology, Beijing, China
}
\begin{document}
%
\maketitle

%
\begin{abstract}
Spatiotemporal prediction aims to generate future sequences by paradigms learned from historical contexts. It is essential in numerous domains, such as traffic flow prediction and weather forecasting. Recently, research in this field has been predominantly driven by deep neural networks based on autoencoder architectures. However, existing methods commonly adopt autoencoder architectures with identical receptive field sizes. To address this issue, we propose an Asymmetric Receptive Field Autoencoder (ARFA) model, which introduces corresponding sizes of receptive field modules tailored to the distinct functionalities of the encoder and decoder. In the encoder, we present a large kernel module for global spatiotemporal feature extraction.  In the decoder, we develop a small kernel module for local spatiotemporal information reconstruction. Experimental results demonstrate that ARFA consistently achieves state-of-the-art performance on popular datasets. Additionally, we construct the RainBench, a large-scale radar echo dataset for precipitation prediction, to address the scarcity of meteorological data in the domain. 
\end{abstract}
\begin{keywords}
Asymmetric receptive field, autoencoder, spatiotemporal prediction
\end{keywords}
\section{Introduction}
\label{intro}

\begin{figure}[tb]
\centering
\includegraphics[width=\linewidth]{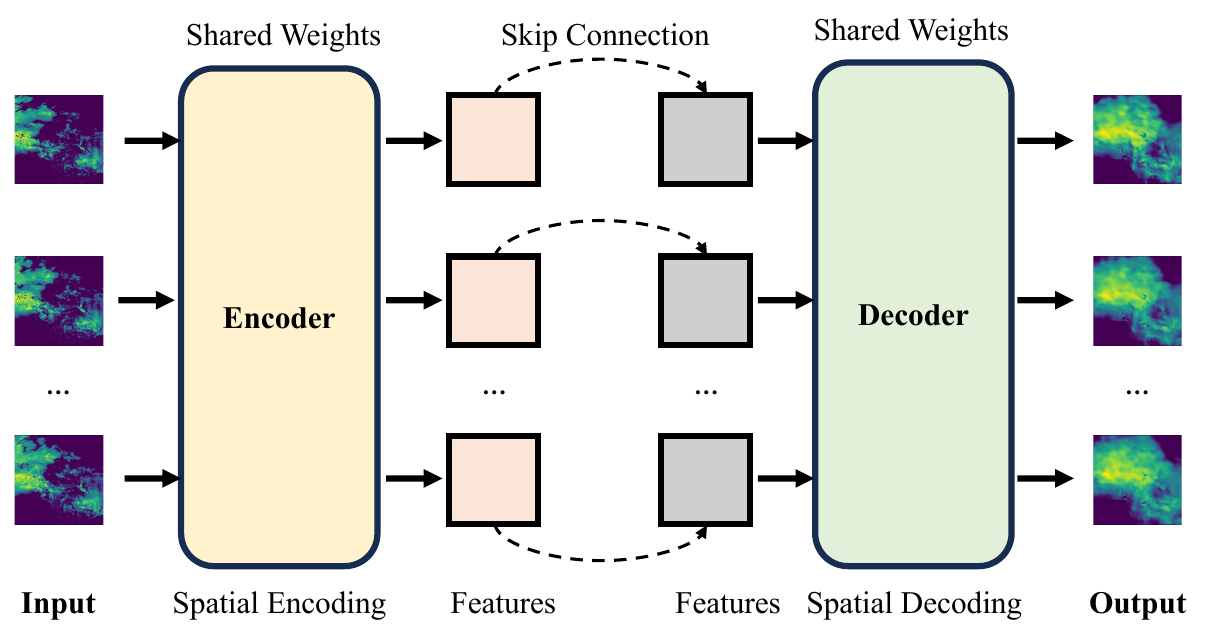}
\caption{Overall pipeline for spatiotemporal prediction using the encoder and decoder with shared weights.}
\label{fig: pipeline}
\end{figure}

\begin{figure*}[t]
\centering
\includegraphics[width=\linewidth]{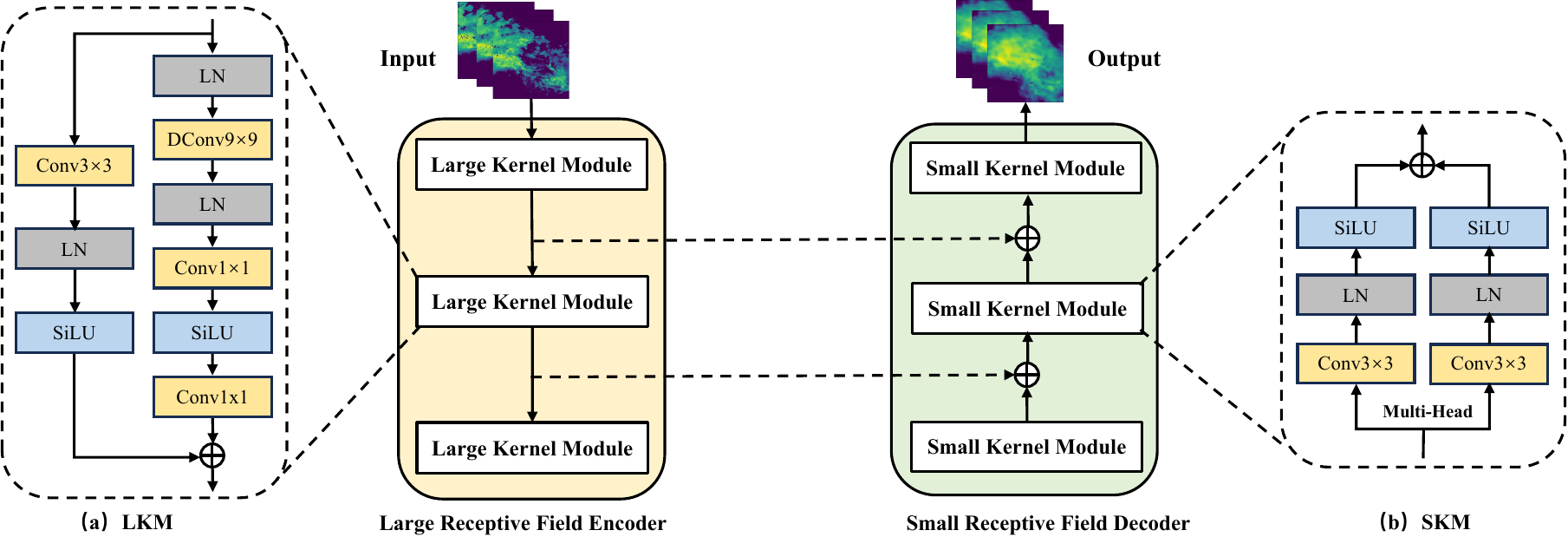}
\caption{Overall architecture of our proposed ARFA. ARFA is an autoencoder consisting of carefully designed Large Kernel Modules (LKM) and Small Kernel Modules (SKM), serving as the encoder and decoder, respectively. The LKMs offer a large receptive field for global feature extraction in the encoder, while the decoder utilizes SKMs for local information reconstruction.}
\label{fig:arfa}
\end{figure*}


Spatiotemporal prediction~\cite{1}, also known as video prediction, aims to synthesize temporal outputs through the input data with temporal relationships. The key is to handle the characteristics of time and space correctly. Recently, PredRNN ~\cite{12,13} has designed a new spatiotemporal LSTM (ST-LSTM)~\cite{21} unit based on RNN ~\cite{19,20,29,30}, which memorizes spatial and temporal features in a unified memory unit and transmits memory at both vertical and horizontal levels. SimVPv2~\cite{48,49} has devised a GSTA structure based on the CNN ~\cite{17,18,diffcr,frcnn}, architecture to effectively capture contextual features in both spatial and temporal dimensions.

Previous models, including both encoder and decoder components, have utilized modules with the same receptive field size without specific optimization, which will reduce the model's accuracy for specific spatiotemporal tasks and increase computational complexity,  as illustrated in Figure~\ref{fig: pipeline}. Additionally, meteorological prediction is one of the most prevalent applications in spatiotemporal prediction, yet there is currently a lack of relevant data to support this research.


This paper proposes a novel Asymmetric Receptive Field Autoencoder (ARFA) model to address these challenges. Additionally, we constructed a large-scale precipitation prediction dataset called RainBench to tackle the missing meteorological prediction data issue.

This work mainly makes the following contributions:

\begin{itemize}
    \item We propose a novel Asymmetric Receptive Field Autoencoder (ARFA) model that combines global encoding and local decoding to achieve high-precision spatiotemporal prediction.
    \item We design two novel modules for the encoder and decoder, the Large Kernel Module and the Small Kernel Module, leveraging the advantages of different-sized convolutional kernels for global spatiotemporal feature extraction and local fine structure reconstruction.
    \item We create a large-scale radar echo dataset for precipitation prediction, named RainBench, with the potential to serve as a benchmark for future spatiotemporal forecasting endeavors.
\end{itemize}

\section{Asymmetric Receptive Field Autoencoder}

As illustrated in Figure~\ref{fig:arfa},
We propose an autoencoder model with an asymmetric receptive field architecture to enhance spatiotemporal prediction accuracy, which includes the encoder and decoder of different sizes of receptive fields.


\subsection{Encoder with Large Receptive Field}

Traditional encoder architectures~\cite{pmaa,tdanet} have limitations in capturing long-range dependency relationships due to their limited receptive field. We propose a novel residual structure called the Large Kernel Module (LKM) for the encoder, enabling it to capture long-range dependencies and extract more comprehensive global contextual information. Specifically, the LKM consists of two parallel branches: the main and residual branches. The main branch primarily comprises a large kernel convolution and a Multi-Layer Perceptron (MLP). The large kernel convolution employs a 9$\times$9 depth-wise separable convolution, which incurs a small increase in parameter count to achieve a larger receptive field. Subsequently, to compensate for the performance degradation caused by the lack of channel interactions in depth-wise separable convolutions~\cite{dsc}, we introduce an MLP composed of two 1$\times$1 convolutions and a SiLU~\cite{silu} to recalibrate the features. Ultimately, the main branch outputs global features.


\begin{equation}\label{equ:encoder_global}
    \mathbf{F}_{global} = \operatorname{MLP}(\operatorname{DConv_{9\times9}}(\mathbf{F}_{in})),
\end{equation}
where $\mathbf{F_{in}} \in \mathbb{R}^{M \times C \times H \times W}$ denotes the input features, $M$ and $N$ denote the lengths of the input and output temporal sequences, respectively, while $C$, $H$, and $W$ represent the number of channels, height, and width of a single image frame. $\operatorname{DConv_{9\times9}}$ refers to depthwise separable convolution, and $\mathbf{F}_{global}$ represents the features outputted by the main branch of LKM. 

On the other hand, the residual branch primarily consists of a convolutional layer with a 3$\times$3 kernel size, which facilitates the extraction of local features. Moreover, layer normalization is introduced to stabilize the training process.

\begin{equation}\label{equ:encoder_global}
    \mathbf{F}_{local} = \sigma(\phi(\operatorname{Conv_{3\times3}}(\mathbf{F}_{in}))),
\end{equation}
where $\sigma$ denotes the activation function, $\phi$ refers to layer normalization, and $\operatorname{Conv_{3\times3}}$ presents standard convolution.

\subsection{Decoder with Small Receptive Field}

 A smaller receptive field in the decoder is advantageous in capturing the local information of the input image more effectively, enabling the decoder to pay greater attention to details and local textures, thereby enhancing the quality of image reconstruction. 
 Motivated by these insights, we have developed a novel Small Kernel Module (SKM), designed to leverage multi-head mechanisms~\cite{24} to augment the decoder's ability for local reconstruction.

Specifically, the SKM is a multi-head structure with two parallel branches. Each branch is composed of a 3$\times$3 convolution, a layer normalization, and a SiLU. Finally, the locally fine-grained features extracted from the two branches are fused through an additive operation to acquire robust features further. The specific formulation is given as follows:

\begin{equation}\label{equ:encoder_global}
    \mathbf{F}_{out} = \sigma(\phi(\mathbf{F}_{global}+\mathbf{F}_{local})),
\end{equation}
where $\mathbf{F}_{out}$ denotes the output features. 




\section{Experiments and Results}

\subsection{Experimental Setup}

\begin{table}
\small
\centering
\caption{A comparative analysis of our RainBench dataset and two prominent spatiotemporal prediction datasets.}
\label{tab:dataset}
\setlength{\tabcolsep}{0pt}
\begin{tabular}{lccc}
\hline
\textbf{Dataset}       & \textbf{Moving-MNIST}~\cite{51} & \textbf{KTH}~\cite{52}            & \textbf{RainBench (Ours)} \\ \hline
\textbf{Train Sample} & 10, 000        & 5, 200           & \textbf{31, 600}            \\
\textbf{Test Sample}  & 10, 000        & 3, 167           & \textbf{31, 600}            \\
\textbf{Frame Size}    & 64$\times$64 & 128$\times$128 & \textbf{200$\times$200}   \\
\textbf{Input Length}  & 10           & 10             & 10               \\
\textbf{Output Length} & 10           & 10             & 10               \\ \hline
\end{tabular}
\end{table}

The dataset and the implementation details are as follows.

\indent\textbf{Dataset.} To address the data scarcity issue in the application of spatiotemporal prediction in meteorology, we have constructed a large-scale radar echo dataset called RainBench for precipitation forecasting. A comparative analysis was conducted with the mainstream spatiotemporal prediction datasets, such as Moving-MNIST~\cite{51} and KTH~\cite{52}, as shown in Table~\ref{tab:dataset}. We collected radar echo precipitation data from the meteorological station in Yinchuan City for the months of April to October 2018 and 2019. This region is characterized not only by low precipitation amounts but also by uneven spatiotemporal distribution of precipitation. The primary precipitation features in this area include infrequent occurrence of rainfall and snowfall, as well as strong water evaporation. Consequently, spatiotemporal prediction in this region poses a significant challenge.

\indent\textbf{Implementation Details.} The batch size for all model training is set to 16, and all experiments are conducted at 200 epochs. We employ four widely adopted metrics to evaluate on the test set: MSE, MAE, PSNR, and SSIM~\cite{56}.



\begin{table}[t]
\centering
\caption{Quantitative comparison of receptive field size.}
\label{tab:rfs}
\setlength{\tabcolsep}{3.5pt}
\begin{tabular}{cccccc}
\hline
\textbf{Encoder} & \textbf{Decoder} & \textbf{MSE}~$\downarrow$            & \textbf{MAE}~$\downarrow$            & \textbf{SSIM}~$\uparrow$          & \textbf{PSNR}~$\uparrow$           \\ \hline
Small   & Small   & 27.102          & 78.379          & 0.939          & 38.280          \\
Small   & Large   & 29.973          & 83.104          & 0.932          & 38.114          \\
Large   & Large   & 28.154          & 78.927          & 0.936          & 38.274          \\
\rowcolor[RGB]{217,217,217} \textbf{Large}   & \textbf{Small}   & \textbf{25.306} & \textbf{73.892} & \textbf{0.944} & \textbf{38.434} \\ \hline
\end{tabular}
\end{table}

\begin{table}[t]
\centering
\caption{Ablation of convolution kernel size in the LKM.}
\label{tab:kernel_size}
\setlength{\tabcolsep}{7pt}
\begin{tabular}{ccccc}
\hline
\textbf{Kernel Size} & \textbf{MSE}~$\downarrow$   & \textbf{MAE}~$\downarrow$   & \textbf{SSIM}~$\uparrow$ & \textbf{PSNR}~$\uparrow$  \\ \hline
3$\times$3           & 26.605 & 76.354 & 0.940 & 38.361 \\
5$\times$5           & 26.102 & 75.495 & 0.941 & 38.383 \\
7$\times$7           & 26.004 & 75.278 & 0.941 & 38.385 \\
\rowcolor[RGB]{217,217,217} \textbf{9$\times$9}           & \textbf{25.975} & \textbf{74.783} & \textbf{0.942} & \textbf{38.398} \\
11$\times$11           & 25.999 & 75.265 & 0.942 & 38.366 \\ \hline
\end{tabular}
\end{table}

\subsection{Ablation and Analysis} 
To validate the effectiveness of our proposed asymmetric receptive field autoencoder model, we conduct four sets of ablation experiments on the most popular Moving-MNIST dataset, and the results are presented in Table~\ref{tab:rfs} and Table~\ref{tab:kernel_size}. 

\textbf{Asymmetric Receptive Field in the Autoencoder.} As shown in Table~\ref{tab:rfs}, in the design of the autoencoder, the optimal combination for achieving the best performance is to employ a large receptive field in the encoder and a small receptive field in the decoder.

\textbf{Convolutional Kernel Size in the LKM.} The size of the convolutional kernel in the LKM directly influences the encoder's receptive field. Table~\ref{tab:kernel_size} demonstrates that as the kernel size increases, the encoder exhibits more effective feature extraction with the support of a larger receptive field. Further increasing the kernel size leads to a performance decline. 

\subsection{Comparison with the State-of-the-Arts}

\begin{table}[t]
\centering
\caption{Quantitative comparison on the Moving-MNIST.}
\label{tab:movingmnist}
\setlength{\tabcolsep}{4pt}
\begin{tabular}{lcccc}
\hline
\textbf{Method}         & \textbf{MSE} $\downarrow$    & \textbf{MAE} $\downarrow$   & \textbf{SSIM} $\uparrow$ & \textbf{PSNR} $\uparrow$  \\ \hline
ConvLSTM~\cite{27}       & 34.071  & 97.250 & 0.919 & 37.614 \\
PhyDNet~\cite{20}        & 28.207  & 78.501 & 0.937 & 38.133 \\
MAU~\cite{19}            & 26.842  & 78.049 & 0.940 & 38.197 \\
PredRNN~\cite{12}        & 33.472  & 95.148 & 0.917 & 37.792 \\
PredRNN++~\cite{13}      & 50.374  & 108.13 & 0.904 & 37.978 \\
E3DLSTM~\cite{29}        & 44.652  & 83.778 & 0.924 & 39.749 \\
PredRNNv2~\cite{31}      & 27.730. & 80.840 & 0.937 & 38.293 \\
SimVPv1~\cite{48}        & 32.203  & 89.291 & 0.927 & 37.951 \\
SimVPv2~\cite{49}        & 27.102  & 78.379 & 0.939 & 38.280 \\
\rowcolor[RGB]{217,217,217} \textbf{ARFA (Ours)}    & \textbf{25.306}  & \textbf{73.892} & \textbf{0.944} & \textbf{38.434} \\
\hline
\end{tabular}
\end{table}

\begin{table}[t]
\centering
\caption{Quantitative comparison on the KTH dataset.}
\label{tab:kth}
\setlength{\tabcolsep}{5pt}
\begin{tabular}{lcccc}
\hline
\textbf{Method}     & \textbf{MSE} $\downarrow$   & \textbf{MAE} $\downarrow$    & \textbf{SSIM} $\uparrow$ & \textbf{PSNR} $\uparrow$  \\ \hline
ConvLSTM~\cite{27}  & 65.164 & 588.333 & 0.855 & 31.215 \\
PredRNN~\cite{12}   & 60.897 & 515.372 & 0.867 & 32.383 \\
SimVPv2~\cite{49}   & 47.541 & 463.063 & 0.903 & 32.740 \\
\rowcolor[RGB]{217,217,217} \textbf{ARFA (Ours)} & \textbf{42.891} & \textbf{403.632} & \textbf{0.907} & \textbf{33.667} \\
\hline
\end{tabular}
\end{table}

\begin{table}[t]
\centering
\caption{Quantitative comparison on the RainBench dataset.}
\label{tab:rainbench}
\setlength{\tabcolsep}{4.5pt}
\begin{tabular}{lcccc}
\hline
\textbf{Method}     & \textbf{MSE} $\downarrow$     & \textbf{MAE} $\downarrow$    & \textbf{SSIM} $\uparrow$ & \textbf{PSNR} $\uparrow$  \\ \hline
ConvLSTM~\cite{27}  & 132.936  & 976.618 & 0.797 & 35.737 \\
PredRNN~\cite{12}   & 128.589  & 877.229 & 0.829 & 36.833 \\
PredRNN++~\cite{13} & 119.671  & 956.249 & 0.795 & 35.710 \\
MIM~\cite{30}       & 119.654  & 874.465 & 0.785 & 34.992 \\
SimVPv2~\cite{49}   & 112.383 & 826.336 & 0.834 & 36.732 \\
\rowcolor[RGB]{217,217,217} \textbf{ARFA (Ours)} & \textbf{109.880}   & \textbf{809.190}  & \textbf{0.842} & \textbf{36.844} \\
\hline
\end{tabular}
\end{table}

\begin{figure*}[t]
\centering
\includegraphics[width=\linewidth]{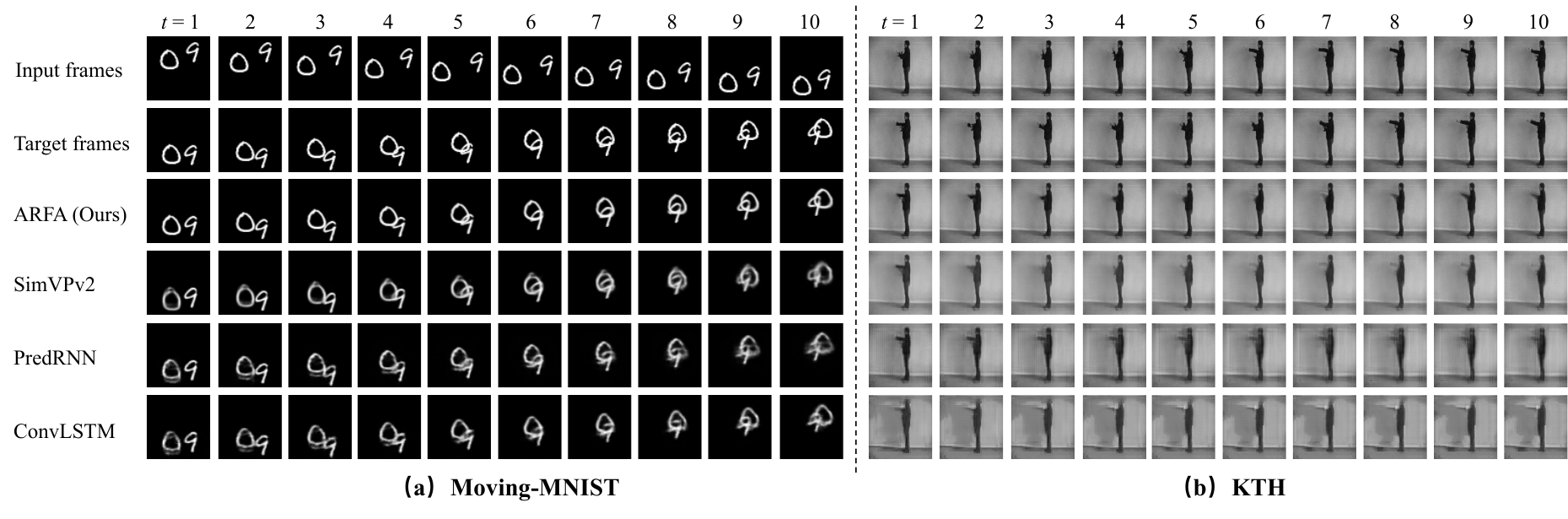}
\caption{Visual results of our ARFA and existing methods on the Moving-MNIST and KTH dataset.}
\label{fig:moving_kth}
\end{figure*}



\begin{figure}[t]
\centering
\includegraphics[width=\linewidth]{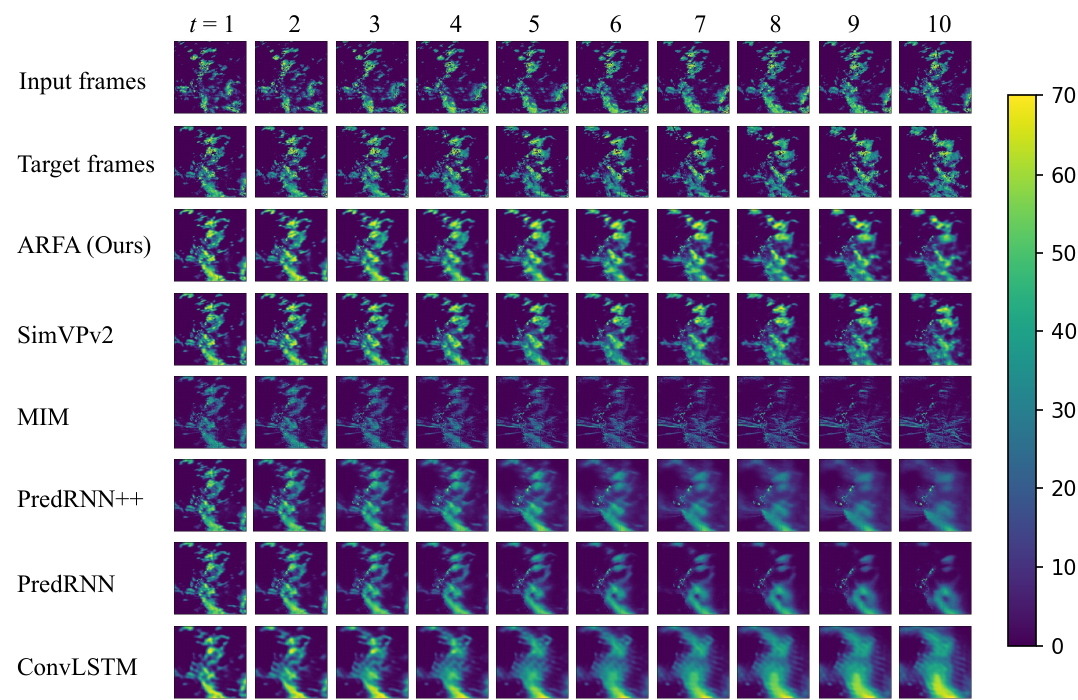}
\caption{A visual comparison on the RainBench dataset.}
\label{fig:rainbench}
\end{figure}

We conduct extensive experiments to quantitatively and qualitatively compare the performance of our proposed method, ARFA, with existing methods for spatiotemporal prediction on two popular datasets, Moving-MNIST and KTH, and on our custom-built RainBench dataset.

\indent\textbf{Moving-MNIST.} Table~\ref{tab:movingmnist} shows the quantitative comparison results of our ARFA and existing approaches on the Moving-MNIST dataset. ARFA achieves state-of-the-art (SOTA) performance consistency across four evaluation metrics. Compared to SimVPv2, ARFA exhibits reductions of 1.796 and 4.487 in MSE and MAE. As shown in Figure~\ref{fig:moving_kth} (a), it can be observed that ARFA consistently generates high-fidelity images. In contrast, other models, such as SimVPv2, PredRNN, and ConvLSTM, suffer from image blurring, while PredRNN exhibits semantic inconsistencies.

\indent\textbf{KTH.} Table~\ref{tab:kth} shows the quantitative comparison results of our proposed ARFA method and existing approaches on the KTH dataset. ARFA achieves SOTA performance in terms of four evaluation metrics. Compared to SimVPv2, ARFA demonstrates a reduction of 4.650 and 59.431 in terms of MSE and MAE. As depicted in Figure~\ref{fig:moving_kth} (b), it can be observed that SimVPv2 fails to restore the arm of the person, and the generated result by  ConvLSTM fails to predict the head and arms. In contrast, images generated by ARFA exhibit no such blurriness or semantic inconsistencies.

\indent\textbf{RainBench.} Table~\ref{tab:rainbench} shows the quantitative comparison results of our ARFA and existing approaches on our RainBench dataset. ARFA achieves SOTA performance consistency across four evaluation metrics. Compared to SimVPv2, 
ARFA reduces the MSE by 2.503 and the MAE by 17.146. Moreover, ARFA exhibits improvements in SSIM and PSNR. As depicted in Figure~\ref{fig:rainbench}, it can be observed that ARFA consistently outperforms other methods in terms of prediction accuracy. Given the inherent challenges of the dataset, future research endeavors could employ RainBench as a benchmark to further explore superior spatiotemporal prediction algorithms. SimVPv2 follows as the second-best performer. However, alternative methods such as MIM, PredRNN, and ConvLSTM gradually lose the original characteristics of the generated images as the prediction time steps increase. 


\section{Conclusion}

This work presents an Asymmetric Receptive Field Autoencoder model to handle spatiotemporal correlations in spatiotemporal prediction. Additionally, RainBench, a large-scale dataset specific to the characteristics of China, is constructed for precipitation prediction. Experimental results confirm the effectiveness of our proposed ARFA. The source code and dataset will be released to facilitate further research.

\section{Acknowledge}
This paper is supported by the National Natural Science Foundation of China (No. 62162053, No. 42265010, No. 62222606, and No. 62076238) and the Natural Science Foundation of Qinghai Province (2023-ZJ-906M).

\vfill\pagebreak



\bibliographystyle{IEEEbib}
\bibliography{refs}

\end{document}